\documentclass[article]{elsarticle}
\usepackage{lineno,hyperref}
\modulolinenumbers[5]

\journal{TBD}

\usepackage{graphicx}
\usepackage{latexsym}
\usepackage{color}
\usepackage{caption}
\usepackage{subcaption}
\usepackage{bm}
\usepackage{bbm}
\usepackage{amsmath,amssymb, amsthm}
\usepackage{algorithm}
\usepackage{algpseudocode}
\usepackage{color,soul}
\usepackage{graphics}
\usepackage{amsfonts}
\usepackage{setspace}   
\usepackage{wasysym}
\usepackage{multirow}

\usepackage{adjustbox}
\usepackage{graphicx}
\usepackage{multirow}
\usepackage{nomencl}
\usepackage{float}
\usepackage[table]{xcolor}
\usepackage{standalone}

\makenomenclature
\biboptions{sort&compress}

\makeatletter
\newcommand{\thickhline}{%
	\noalign {\ifnum 0=`}\fi \hrule height 1pt
	\futurelet \reserved@a \@xhline
}

\makeatletter
\def\ps@pprintTitle{%
	\let\@oddhead\@empty
	\let\@evenhead\@empty
	\def\@oddfoot{\reset@font\hfil\thepage\hfil}
	\let\@evenfoot\@oddfoot
}
\makeatother

\newcolumntype{"}{@{\hskip\tabcolsep\vrule width 1pt\hskip\tabcolsep}}
\makeatother

\newcolumntype{M}[1]{>{\centering\arraybackslash}m{#1}}

\usepackage{geometry}
\begin{document}

\begin{frontmatter}

\title{Physics-Driven Regularization of Deep Neural Networks for~Enhanced Engineering Design and Analysis}

\author{Mohammad Amin Nabian}
\author{Hadi Meidani\corref{mycorrespondingauthor}}
\address{Department of Civil and Environmental Engineering, University of Illinois at Urbana-Champaign, Urbana, Illinois, USA.}
\cortext[mycorrespondingauthor]{Corresponding author}
\ead{meidani@illinois.edu}

\begin{abstract}
	In this paper, we introduce a physics-driven regularization method for training of deep neural networks (DNNs) for use in engineering design and analysis problems. In particular, we focus on prediction of a physical  system, for which in addition to  training data, partial or complete information on a set of governing laws is also available. These laws often appear in the form of differential equations, derived from first principles, empirically-validated laws, or domain expertise, and are usually neglected in data-driven prediction of engineering systems. We propose a training approach that utilizes the known governing laws and regularizes data-driven DNN models by penalizing divergence from those laws. The first two numerical examples are synthetic examples, where we  show that in constructing a DNN model that best fits the measurements from a physical system, the use of our proposed regularization results in DNNs that are more  interpretable with smaller generalization errors, compared to other common regularization methods. The last two examples concern metamodeling for a random Burgers' system and for aerodynamic analysis of passenger vehicles, where we demonstrate that the proposed regularization provides superior generalization accuracy compared to other common alternatives.
\end{abstract}


\end{frontmatter}


\section{Introduction} \label{sec:intro}

Deep neural networks (DNNs) have attracted tremendous attention in various research fields in engineering disciplines. These models usually consist of a large number of parameters and  as such are extremely flexible. This is beneficial for learning, but also highly challenging when  overfitting is concerned. That is, even though the model fits very well to  training data, it cannot effectively find the underlying relationship in data and as a result, DNNs may not generalize well to  unobserved test data. In order to overcome this difficulty, several regularization techniques are developed to prevent DNNs from overfitting and improve the generalization accuracy \cite{goodfellow2016deep}. This is done particularly during the training, where the regularizers  incorporate  a penalty term  into the loss function that is to be minimized during model training. Current popular choices of regularization methods for DNNs include $L^{2}$ and $L^{1}$ regularizations \cite{lecun2015deep}, and dropout \cite{srivastava2014dropout}.

This paper presents an approach for physics-driven regularization of DNNs for systems that are subject to known governing laws in form of a partial differential equation (PDE). These governing laws are  obtained using either  first principles, empirically-validated laws, and/or  domain expertise. In data-driven modeling of physical systems, this prior knowledge is usually available, but not directly used in training of these models (e.g. \cite{guo2016convolutional,hennigh2017lat,hennigh2017automated}). In training of DNNs, in particular, the prior knowledge about the governing equations can be effectively utilized to ``push" the trained models to satisfy the governing laws. In this work, we do so by creating a regularization term that accounts for the underlying physics; a term that penalizes divergence form the governing equations. It is shown through numerical examples that this regularization method offers two main advantages: (1) it effectively prevents overfitting and results in significantly smaller generalization errors, when compared to other  regularization methods; and (2) it produces models that are more physically interpretable, as opposed to the ones that are trained using purely data-driven approaches.

In the past few decades, several machine learning methods have been used as an approximate solution for physical or biological systems. Examples include polynomial response surfaces (e.g. \cite{queipo2005surrogate,shanock2010polynomial}), radial basis functions (e.g. \cite{buhmann2000radial,wild2008orbit}), polynomial chaos expansions (e.g. \cite{xiu2002wiener,marzouk2009dimensionality}), Kriging (e.g. \cite{simpson2001kriging,jeong2005efficient}), Gradient-Enhanced Kriging (GEK) (e.g. \cite{bouhlel2017gradient,de2012efficient}), and Support Vector Regression (SVR) (e.g. \cite{smola2004tutorial,clarke2005analysis}). Examples of other machine-learning studies for data-driven discovery of PDEs include Ridge regression (e.g. \cite{rudy2017data}), Gaussian process regression (e.g. \cite{raissi2018hiddenb}), polynomial approximation \cite{wu2019numerical}, and nonparametric nonlinear regression (e.g. \cite{voss1999amplitude}). Another machine learning method which has been successfully used is the DNN.

When applied to physical systems, DNN works can be categorized into different groups, based on the specific learning task they address and also based on what they exploit, (empirical or simulation) data and/or underlying physics or governing laws. A non-exhaustive list of the tasks is presented in Table \ref{table:literature}. The first group includes works which use no information on the physics of the system, because  it is either unavailable or too complex. The tasks in this group include data-driven modeling and metamodeling (e.g. \cite{shan2018study,nabian2018deep1}), and data-driven discovery of PDEs (e.g. \cite{qin2018data}). The studies in the second and third category utilize the available information on the physics of the system, and are generally inspired by the methodology presented in \cite{lagaris1998artificial} for solving differential equations, where the authors used neural networks as a trial solution to these equations, and trained this trial solution such that it satisfies the differential equations.

The second group of works, where DNNs are trained exclusively based on the known governing equations, includes unsupervised learning tasks for solving deterministic and random PDEs. For instance, in \cite{lagaris1998artificial} neural networks are used to solve initial and boundary value problems, where the neural network parameters are  calibrated by minimizing the squared residuals over specified collocation points. A similar approach was used in \cite{raissi2019physics} to solve more challenging dynamic problems described by time-dependent nonlinear PDEs. A mesh-less variation of these approaches is introduced in \cite{sirignano2018dgm} for solving high-dimensional PDEs with applications in finance. Also in \cite{nabian2018deep2}, deep residual learning is used for solving high-dimensional random PDEs in strong and variational forms. The deep learning methods in this group can be directly compared with traditional numerical methods for solving deterministic PDEs (such as finite difference, finite element, or finite volume methods), or random PDEs (such as Stochastic Collocation \cite{xiu2005high}, Stochastic Galerkin \cite{ghanem2003stochastic}, or Monte Carlo methods \cite{fishman2013monte}), versus which  they   offer  advantages.

In the third group, both the data and partial or complete information about the underlying physics are needed in the learning task.  Examples in this group include \cite{raissi2019deep,raissi2018hiddena}, where the task at hand  is data assimilation, which is  not solvable by a pure machine learning approach, that is if knowledge about the physics is not exploited. Particularly in these two works, for a system governed by the Navier-Stokes equations, DNNs are used for inference about a latent quantity, e.g. the lift and drag forces on structures exposed to a fluid flow, given indirect observation/simulation data on a flow feature, e.g. the velocity field. Such data assimilation task addresses the case where a conventional supervised machine learning approach or a pure scientific computing approach cannot be used independently \cite{raissi2018hiddena}.

In this paper, we focus on a task in the third group, and introduce a physics-driven regularization method that can be used in presence of simulation or measurement data to train more \emph{accurate} and \emph{interpretable} DNN regression models for engineering analysis and design optimization. Precisely, in this study we focus on building regression models, and numerically show that although a pure data-driven deep learning approach can address this problem (as opposed to, e.g., problems in \cite{raissi2019deep,raissi2018hiddena}), when this learning approach is used with our proposed regularization, the resulting model can have better generalization accuracy and interpretability. To demonstrate this, in particular, we consider systems governed by the Burgers' and Navier-Stokes equations with a variety of setups for initial/boundary conditions, and provide a thorough comparison between the performance of the model regularized by our method, and those regularized by other alternatives commonly used in the literature, in presence of measurement/simulation data with different noise levels.

The remainder of this paper is organized as follows. Feed-forward fully-connected DNNs are explained in Section \ref{sec:DNN}. In Section \ref{sec:Reg}, a number of commonly-used regularization methods for training of DNNs are discussed. Our proposed regularization method is then introduced in Section \ref{sec:PhysReg}. Section \ref{sec:examples} presents four numerical examples, on which the performance of the proposed regularization is compared with other common alternatives. Finally,  Section \ref{sec:conclusion} concludes the paper, with some discussion on the relative advantages and limitations of this regularization method and also potential future works.

\begin{table*}[]
	\begin{center}
		\caption{A non-exhaustive list of deep-learning tasks introduced in the literature for modeling of physical systems.}
		\label{table:literature}
		\scalebox{0.9}{
			{\begin{tabular}{|c|l|}
					\hline
					Use of data or physics in training & Tasks                          \\ \hline \hline
					Only data & \begin{tabular}[l]{@{}l@{}}Data-driven modeling and metamodeling (e.g. \cite{lee1990neural,wang1990structured,mills2017deep,ehrhardt2017learning,shan2018study,zhang2018solving})
						\\ Data-driven discovery of PDEs (e.g. \cite{raissi2019physics,qin2018data,gonzalez1998identification})   \end{tabular} \\ \hline
					Only physics & Unsupervised learning for solving PDEs (e.g. \cite{lagaris1998artificial,raissi2019physics,sirignano2018dgm,nabian2018deep2,rudd2013solving,weinan2017deep,han2018solving,berg2018unified})    \\ \hline
					Both data and physics & \begin{tabular}[l]{@{}l@{}}Physics-informed data assimilation (e.g. \cite{raissi2019deep,raissi2018hiddena})\\ Regression with physics-driven regularization (present study)    \end{tabular}   \\ \hline
		\end{tabular}}	}
	\end{center}
\end{table*}

\section{Regression using feed-forward fully-connected DNNs} \label{sec:DNN}

In a regression task, the objective is to approximate an unknown function ${u}$ given a training dataset consisting of $n$ input samples $\bm{X}=\{\bm{x}_1,\cdots,\bm{x}_n\}$, $\bm{X} \in \mathbb{R}^{n \times d}$ and their corresponding outputs $\bm{Y}=\{\bm{y}_1,\cdots,\bm{y}_n\}$, $\bm{Y} \in \mathbb{R}^{n \times k}$. Specifically, we consider the following relationship holds for any data $\left(\bm{x}_i,\bm{y}_i\right)$, $\forall i \in \{1, \cdots, n\}$
\begin{equation}
\bm{y}_i = u\left(\bm{x}_i\right) + \epsilon_i,
\end{equation}
where ${u}(\cdot)$ is the unknown nonlinear function and $\epsilon_i$ is the measurement or simulation noise. In our case, $\bm{y}_i$  is either the solution of a PDE given the input variables $\bm{x}_i$, such as time and/or spatial coordinates, or the measurement from a system governed by a PDE. Our objective is to approximate the function $u$ by a DNN. 

For notation brevity, let us first define the \textit{single hidden layer} neural networks, since the generalization of a single hidden layer network to a network with multiple hidden layers, effectively a \emph{deep} neural network, will be straightforward. Specifically, given an input $\bm{x}_i \in D^{d}$, a standard single hidden layer neural network approximates the $k$-dimensional response according to \begin{equation} \label{OHL-NN}
\bm{y}_i \approx \tilde{u}\left(\bm{x}_i;\bm{\Theta}\right)  = \sigma (\bm{x}_i \bm{W}_{1}+\bm{b}_{1}   ) \bm{W}_{2}+\bm{b}_{2},
\end{equation}
where $\bm{W}_{1}$ and $\bm{W}_{2}$ are weight matrices of size $d\times q$ and $q\times k$, and $\bm{b}_{1}$ and $\bm{b}_{2}$ are \emph{bias} vectors of size $1\times q$ and $1\times k$, respectively. The function $\sigma( \cdot  )$ is an element-wise non-linear model, commonly known as the \textit{activation} function. Popular choices of activation functions include Sigmoid, hyperbolic tangent (Tanh), and Rectified Linear Unit (ReLU). In deep neural networks, the output of each activation function is transformed by a new weight matrix and a new bias, and is then fed to another activation function. Each new set of weight matrices and biases added to Equation (\ref{OHL-NN}) constitutes a new \textit{hidden layer} in the neural network. Generally, the capability of neural networks to approximate complex nonlinear functions can be improved by adding more hidden layers or increasing the dimensionality of the hidden layers \cite{lecun2015deep,goodfellow2016deep}. 

In order to calibrate the weight matrices and biases, we use a Euclidean loss function as follows
\begin{equation} \label{MSE Loss}
J(\bm{\Theta};  \bm{X},\bm{Y})=\frac{1}{2n}\sum_{i=1}^{n}\left \| \bm{y}_i- \tilde{u}\left(\bm{x}_i;\bm{\Theta}\right) \right \|^{2},
\end{equation}
where $J(\cdot)$ is the loss function, $\bm{\Theta}=\{\bm{W},\bm{b} \}$, $\bm{W}=\{\bm{W_1},\bm{W_2},\cdots \}$, $\bm{b}=\left\{ \bm{b_1},\bm{b_2},\cdots \right\}$. 
The model parameters can be calibrated according to the following optimization problem
\begin{equation} \label{minimize_loss}
\bm{\Theta}^*=\underset{{ \bm{\Theta} }}{\operatorname{argmin}} \; J(\bm{\Theta}; \bm{X},\bm{Y}),
\end{equation}
where $\bm{\Theta}^*$ are the estimated parameter values at the end of the training. The optimization defined in Equation (\ref{minimize_loss}) is performed iteratively using Stochastic Gradient Descent (SGD) and its variants \cite{bottou2012stochastic,kingma2014adam,duchi2011adaptive,zeiler2012adadelta,sutskever2013importance}. Specifically, at the $i^{th}$ iteration, the model parameters are updated according to
\begin{equation} \label{descent step}
\bm{\Theta}^{(i+1)} = \bm{\Theta}^{(i)} - \eta^{(i)} \nabla_{\bm{\Theta}}J^{(i)}(\bm{\Theta}^{(i)}; \bm{X},\bm{Y}),
\end{equation}
where $\eta^{(i)}$ is the step size in the $i^{\textit{th}}$ iteration. At each iteration, $\nabla_{\bm{\Theta}} J^{(i)}(\bm{\Theta}^{(i)}; \bm{X},\bm{Y})$ is calculated using \emph{backpropagation} \cite{lecun2015deep}, where the gradients of the objective function with respect to the weights and biases of a DNN are calculated by starting off from the network output and propagating towards the input layer while calculating the gradients, layer by layer, using the chain rule. More details on the feed-forward fully-connected DNNs can be found in \cite{lecun2015deep,goodfellow2016deep}. 

\section{{Regularization of DNNs}} \label{sec:Reg}
In this section a number of regularization methods for DNNs are briefly introduced, including parameter norm regularizations and dropout, which are the commonly-used methods for DNN regularization among the others (e.g. dropconnect \cite{wan2013regularization}, early stopping \cite{caruana2001overfitting}, and data augmentation \cite{salamon2017deep}).

\subsection{Parameter norm regularization}
Most of the regularization methods limit the flexibility of DNN models by adding a parameter norm penalty term $\Omega(\bm{\Theta})$ to the loss function $J(\bm{\Theta}; \bm{X},\bm{Y})$. The regularized loss function denoted by $\hat{J}(\bm{\Theta}; \bm{X},\bm{Y})$ can be expressed as
\begin{equation}
\hat{J}(\bm{\Theta}; \bm{X},\bm{Y})=J(\bm{\Theta}; \bm{X},\bm{Y})+\lambda\Omega(\bm{\Theta}),
\end{equation}
where $\lambda \in [0,\infty)$ is a hyperparameter controlling the contribution of the parameter norm penalty term relative to the standard loss function $J(\bm{\Theta}; \bm{X},\bm{Y})$. $L^2$ and $L^1$ regularizations are among the most common parameter norm regularizations. We note that for DNNs, parameter norm regularization usually penalizes only on the weights, and biases will remain unregularized, as discussed in \cite{goodfellow2016deep}.

The $L^2 $ parameter regularization is performed by setting the penalty term $\Omega(\bm{\Theta})=\frac{1}{2}||\bm{W}||^2_2$. The $L^2$-regularized loss function $J_{L^2}$ therefore takes the following form
\begin{equation}
J_{L^2}(\bm{\Theta}; \bm{X},\bm{Y})=J(\bm{\Theta}; \bm{X},\bm{Y})+\frac{\lambda_2}{2}\bm{W}^T\bm{W}.
\end{equation}
The addition of the weight decay term has modified the learning rule to multiplicatively shrink the weights by a constant factor on each training iteration. Therefore, $L^2$ regularization forces the DNN parameters toward taking relatively small values.

The $L^1 $ parameter regularization consists of setting the penalty term $\Omega(\bm{\Theta})=\frac{1}{2}||\bm{W}||_1=\sum_{i}|w_i|$, where $\{w_i\}$ are the individual weight parameters of the DNN. The  $L^1$-regularized loss function $J_{L^1}$ therefore takes the following form
\begin{equation}
J_{L^1}(\bm{\Theta}; \bm{X},\bm{Y})=J(\bm{\Theta}; \bm{X},\bm{Y})+\lambda_1 ||{\bm{W}}||_1.
\end{equation}
In comparison to $L^2$ regularization, the $L^1$ regularization contribution to the loss gradient no longer scales linearly with $\bm{W}$ but instead it is a constant. As a result, a regularization is created that effectively promotes sparsity for the weight matrix $\bm{W}$.

\subsection{Dropout}

Dropout involves removing  components of each layer randomly   with probability $P$   during model optimization and for each forward-backward pass, i.e. each iteration to update the model parameters \cite{srivastava2014dropout}. This prevents units from excessive co-adapting \cite{hinton2012improving}. The dropped-out components will not have a contribution to the forward pass and weight updates will not be applied to these components on the backward pass. As a result of applying dropout, effectively an exponential number of different \emph{thinned} networks are sampled. At test time, a single \emph{unthinned} network is used (including all the units) by averaging the predictions of all these thinned networks \cite{srivastava2014dropout}.

The standard single hidden layer neural network defined in Equation \ref{OHL-NN} with dropout takes the following form
\begin{equation} \label{OHL-NN-Dropout}
\bm{y} = \bm{r} \cdot \sigma (\bm{x} \bm{W}_{1}+\bm{b}_{1}   ) \bm{W}_{2}+\bm{b}_{2},
\end{equation}
where $\bm{r}=\{r_1,\cdots,r_k\}$ is a $1 \times k$ vector, and $r_j \sim \text{Bernoulli} (P), \forall j \in \{1,\cdots, k\}$. Dropout is shown to improve the performance of DNNs in a variety of supervised learning tasks in speech recognition, vision, document classification, and computational biology \cite{srivastava2014dropout,hinton2012improving,dahl2013improving,krizhevsky2012imagenet,pham2014dropout}. It is shown in \cite{wager2013dropout} that dropout applied to linear regression is equivalent to $L^2$ regularization.

\section{Physics-Driven Regularization} \label{sec:PhysReg}
As stated earlier in Section \ref{sec:DNN}, in DNN regression we seek to approximate  the unknown response function ${u}$. We consider cases where the response function follows a governing law, as follows
\begin{equation}
\mathcal{N}(\bm{x}_i,u\left(\bm{x}_i\right))  =0, \; \; \; \;  \bm{x}_i\in \mathbb{R}^{d},
\end{equation}
where $\mathcal{N}(\cdot)$ is a general differential operator that may consist of partial derivatives and linear and nonlinear terms. Let us denote the DNN approximation  by $\tilde{u}\left(\bm{x};\bm{\Theta}\right)$. The physics-driven loss function $J_{\text{PD}}\left(\bm{\Theta}; \bm{X},\bm{Y},\mathcal{N}\right)$ is then defined as follows
\begin{equation}\label{PI loss}
J_{\text{PD}}\left(\bm{\Theta}; \bm{X},\bm{Y},\mathcal{N}\right)=J\left(\bm{\Theta}; \bm{X},\bm{Y}\right)+ \lambda_{\mathcal{N}} J_{\mathcal{N}}\left(\bm{\Theta}; \bm{X} \right),
\end{equation}
where $\lambda_{\mathcal{N}} \in [0,\infty)$ is a hyperparameter controlling the contribution of the penalty term   $J_{\mathcal{N}}\left(\bm{\Theta}; \bm{X},\mathcal{N}\right)$ that is defined as
\begin{equation}\label{PI term}
J_{\mathcal{N}}\left(\bm{\Theta}; \bm{X} \right) =\frac{1}{2n}\sum_{i=1}^{n}\left[\mathcal{N}\left(\bm{x}_i,\tilde{u}\left(\bm{x}_i;\bm{\Theta}\right)\right)\right] ^2,
\end{equation}
in which the term $\mathcal{N}(\bm{x}_i,\tilde{u}\left(\bm{x}_i;\bm{\Theta}\right))$ measures the divergence of the DNN solution $\tilde{u}\left(\bm{x}_i;\bm{\Theta}\right)$ from the governing laws at input location $\bm{x}_i$. By adding this regularization term to the standard loss function, the  learning task considers both the mean squared differences between model prediction and measurements (as reflected in $J\left(\bm{\Theta}; \bm{X},\bm{Y}\right)$), the divergence from the governing laws (as reflected in $J_{\mathcal{N}}\left(\bm{\Theta}; \bm{X} \right)$). The term $J_{\mathcal{N}}\left(\bm{\Theta}; \bm{X} \right)$ usually consists of partial derivatives, and in training of DNNs with physics-driven regularization, these derivatives can be calculated using automatic differentiation \cite{baydin2018automatic,griewank2008evaluating}. In automatic differentiation, the gradient is calculated by successively applying the chain rule to break down the operation into a sequence of simple operations and calculating the gradient of each simple operation analytically. It is important to note that automatic differentiation is different from symbolic or numerical differentiation, which are two common alternatives for computing derivatives \cite{baydin2018automatic}.

It is shown through numerical examples that the proposed regularization method effectively prevents DNNs from overfitting, and also results in models that are better physically interpretable. That is, it can estimate more accurately the partial derivatives of the response which carry physical interpretation and can be utilized in  subsequent calculations, such as sensitivity analysis. Although the regularization term shown in Equation (\ref{PI term}) is written over the same inputs as those the standard loss function $J(\bm{\Theta}; \bm{X},\bm{Y})$ is written over, it should be noted that this is not a requirement. As an alternative, especially in problems with lack of sufficient labeled input data, a different and possibly larger set of input data  may be used for this penalty term. Additionally, training the DNN parameters may be performed in a sequential fashion by using the standard loss function first and the regularized loss function at a later stage in training. It is also worthwhile mentioning that the proposed regularization method can generally be combined with other regularization methods. For example, we can use hybrid physics-driven-$L^1$ regularization in order to push the DNN to satisfy governing laws and at the same time promote model sparsity. 

\section{Numerical examples} \label{sec:examples}

In this section, we numerically study the performance of the proposed regularization method in constructing accurate DNN models and metamodels for systems governed by physical laws. In the first and second examples, we consider regression problems when response measurements are obtained from systems that are governed by the Burgers' and Navier-Stokes equations, respectively. In both examples, comparative results are reported for different regularization approaches. In the third example, we show the effectiveness of the proposed regularization method in constructing metamodels for physical systems with random initial and boundary conditions. Finally, in the last example, we construct a DNN metamodel using the proposed regularization  that can be used for vehicle aerodynamic optimization, and show that our proposed method results in smaller generalization error compared to the current state of the practice.

\subsection{DNN regression on response measurements from Burgers' equation}\label{example1}
\begin{figure}
	\begin{center}
		\includegraphics[width=0.55\linewidth]{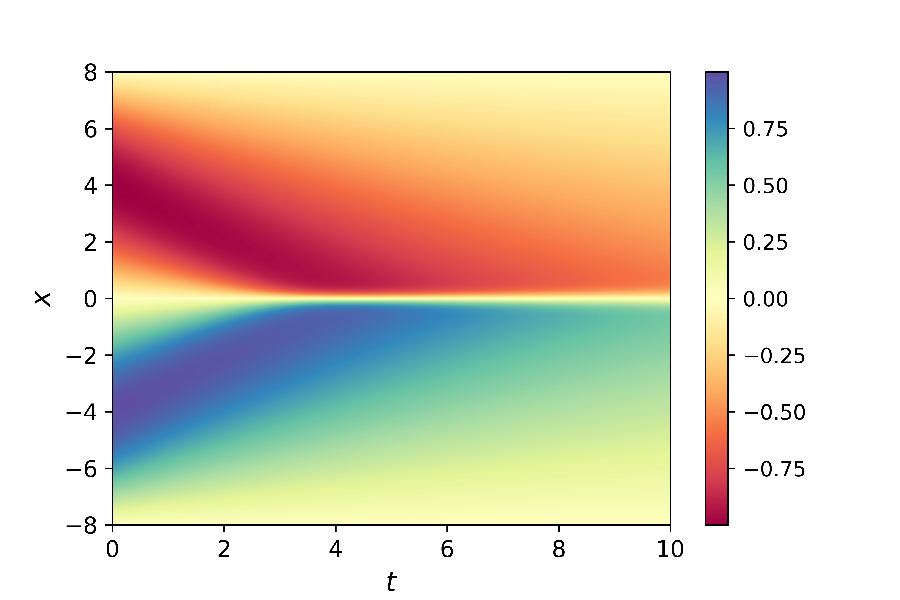}
		\caption{Numerical solution to the Burgers' equation defined in Equation \ref{Eqn:Burgers} (the figure is adapted from \cite{raissi2018deep,raissi2019physics}).} 
		\label{Burgers_snapshot}
	\end{center}
\end{figure}

Systems governed by the Burgers' equation   arise in various areas of engineering, such as traffic flow, fluid mechanics, and acoustics. Let us consider a system governed by the Burger's equation, and that we have full knowledge about the parameter values of the governing equation, but not  the boundary or initial conditions. Let us further assume that we have access to a number of noisy measurements from this system, based on which we seek to construct a DNN regression model. These response measurements could be obtained experimentally from the system of interest, but in this work we generate synthetic measurements by numerically solving the following Burgers' equation, using the source code provided by \cite{raissi2018deep}
\begin{equation}\label{Eqn:Burgers}
\frac{\partial u}{\partial t}+u \frac{\partial u}{\partial x} - 0.1 \frac{\partial ^2u}{\partial x^2}=0,\; \; \; \; \;  
u(0,x)=-\sin(\pi x/8),
\end{equation}
with periodic boundary condition, and $x \in [-8,8], \  t \in [0,10]$. Specifically, in \cite{raissi2018deep}, the Chebfun package \cite{driscoll2014chebfun} is used with a spectral Fourier discretization with 256 modes and a $4^{th}$-order explicit Runge-Kutta scheme, where the size of time steps is set to $10^{-4}$, and the solution is saved every 0.05 s time interval. Using the specifications for the time and space domains used in \cite{raissi2018deep}, the Burgers' equation is solved and the solution at different points in time and space is depicted in Figure \ref{Burgers_snapshot}. Next, we add a Gaussian noise to the solution $u$, with a zero mean and a standard deviation of $\gamma \bar{u}$, where $\bar{u}$ is the mean value of $u$ in the solution dataset, and $\gamma$ is a constant which controls the noise level. In this example we consider three different noise levels, with $\gamma=0$, $\gamma=0.15$, and $\gamma=0.25$.

From the noisy solution dataset, we randomly select 500 samples as our synthetic measurement data, and we assume this is the only available measurement data for training our models. We construct five different DNN models with five different regularization choices:  no regularization, $L^2$ regularization, $L^1$ regularization, dropout, and the proposed regularization. The architecture of these DNN models is fixed and consists of 4 hidden layers, each with 32 units with Tanh nonlinearities. The Adam optimization algorithm \cite{kingma2014adam} is used to solve the optimization problem defined in Equation \ref{minimize_loss}. Parameters $\beta_1$, $\beta_2$, and $\epsilon$ for the Adam optimizer are set to 0.9, 0.999, and $10^{-8}$, respectively. Batch size is set to 50. For the PD regularization, the following penalty term is used
\begin{equation}\label{Burgers_penalty}
J_{\mathcal{N}}\left(\bm{\Theta}; \bm{X}\right)= \frac{1}{2n}\sum_{i=1}^{n}\biggl[\frac{\partial \tilde{u}}{\partial t} \left(\bm{x}_i;\bm{\Theta}\right)+\tilde{u} \left(\bm{x}_i;\bm{\Theta}\right) \frac{\partial \tilde{u}}{\partial x} \left(\bm{x}_i;\bm{\Theta}\right)  -  0.1 \frac{\partial ^2\tilde{u}}{\partial x^2} \left(\bm{x}_i;\bm{\Theta}\right)\biggr] ^2.
\end{equation}
Automatic differentiation is used to calculate partial derivatives of the DNN model with respect to input parameters.

Table \ref{table: example1} shows the hyperparameters that are tuned for each of the models, together with the search domain for each of the hyperparameters. Training is performed for 8 different number of epochs starting from 25,000 epochs and ending with 200,000 epochs. For each regularization method, given the number of epochs, we train 100 models on the training dataset. The model which results in the lowest relative $L_2$ norm on the evaluation dataset is then selected as the best model for the given number of training epochs. In order to eliminate the dependency of the reported relative $ L_2$ error norms to the selection of evaluation and test datasets, we choose to use a rather large number of our available noisy synthetic data, and therefore we use 5000 randomly selected samples for evaluation and test purposes. 

\begin{table*}
	\begin{center}
		\caption{The hyperparameters that are tuned for each of the models together with their search domain.}
		\label{table: example1}
		\scalebox{0.74}{
			\begin{tabular}{|M{2.85cm}||M{2.95cm}|M{2.95cm}|M{2.95cm}|M{2.95cm}|M{2.95cm}|}
				\hline
				Regularization method & No Reg. & $L^2$ Reg. & $L^1$ Reg. & Dropout & PD Reg. \\ \hline
				Hyperparameters & $\eta \in [10^{-6},10^{-4}] $ & $\eta \in [10^{-6},10^{-4}] $ $\lambda_2 \in [10^{-6},10^{-2}]$ & $\eta \in [10^{-6},10^{-4}] $ $\lambda_1 \in [10^{-6},10^{-2}]$ & $\eta \in [10^{-6},10^{-4}] $ $P \in [0.9,0.999]$ & $\eta \in [10^{-6},10^{-4}] $ $\lambda_{\mathcal{N}} \in [10^{-3},10^{1}]$                 \\ \hline
		\end{tabular}}
	\end{center}
\end{table*}

\begin{figure}
	\begin{center}
		\begin{subfigure}[t]{0.42 \linewidth}
			\includegraphics[width=.99\linewidth]{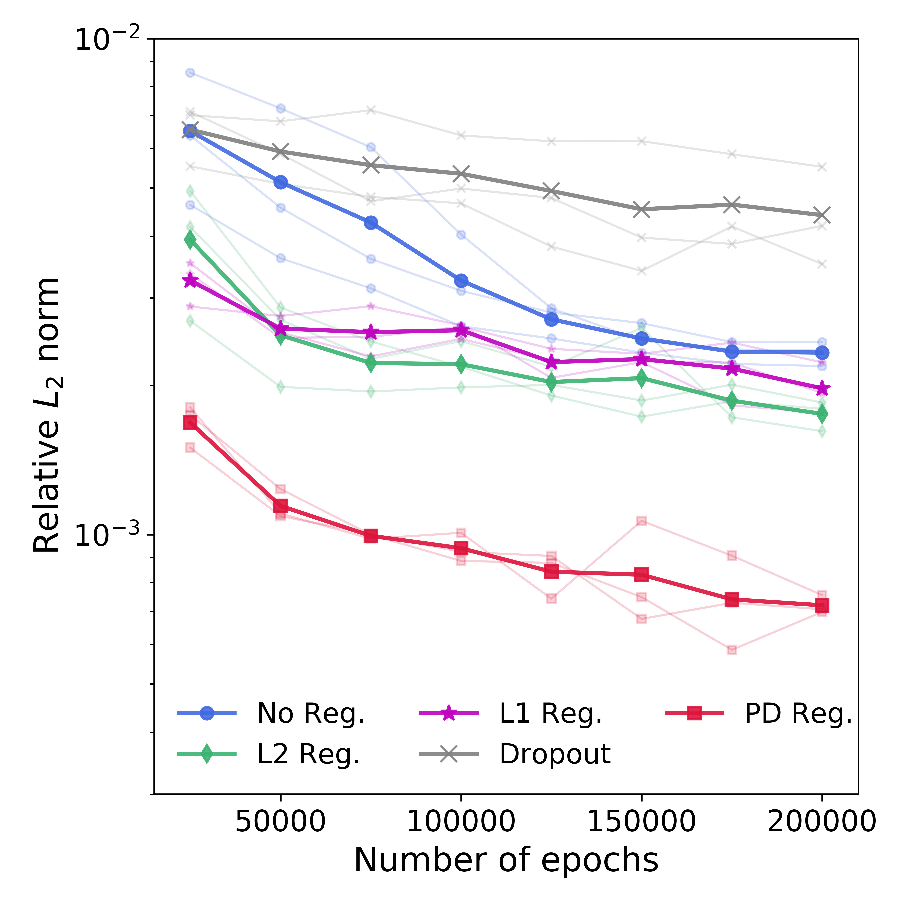}
			\caption{}
			\centering	
		\end{subfigure}
		\quad
		\begin{subfigure}[t]{0.42 \linewidth}
			\includegraphics[width=.99\linewidth]{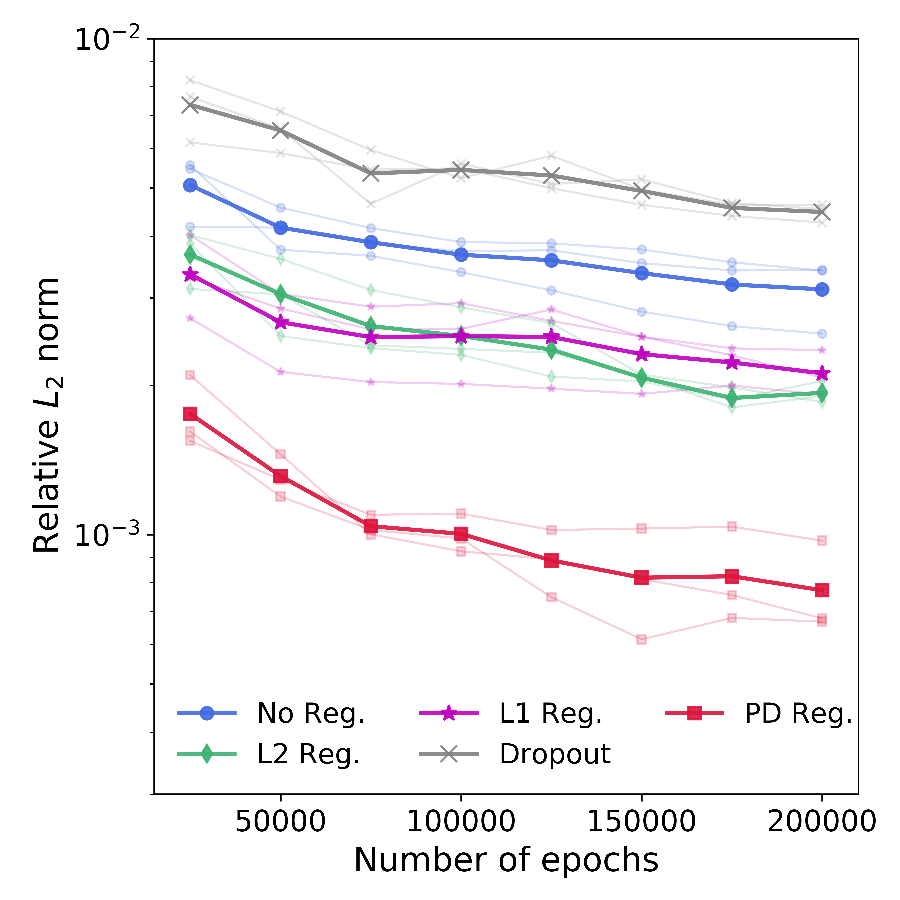}
			\caption{}
		\end{subfigure}
		\\
		\begin{subfigure}[t]{0.42 \linewidth}
			\includegraphics[width=.99\linewidth]{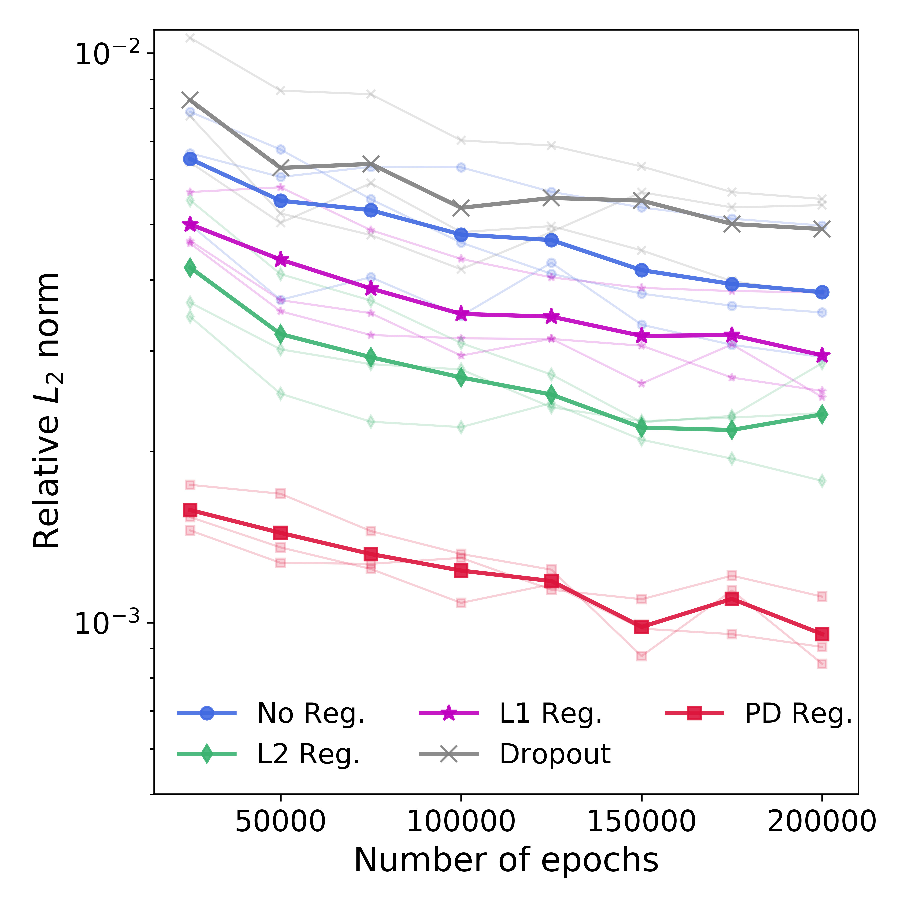}
			\caption{}	
		\end{subfigure}
		\captionsetup{}
		\caption{Comparison between the performance of different regularization methods in terms of relative $L_2$ norm for the Burgers' example: (a) $\gamma=0$; (b) $\gamma=0.15$; (c) $\gamma=0.25$. For each method, the three lines with lighter shades refer to the three different trainings, with the darker shade referring to the average performance of the three trained models. } 
		\label{Burgers_performance}
	\end{center}
\end{figure}

\begin{figure}
	\begin{center}
		\includegraphics[width=.67\linewidth]{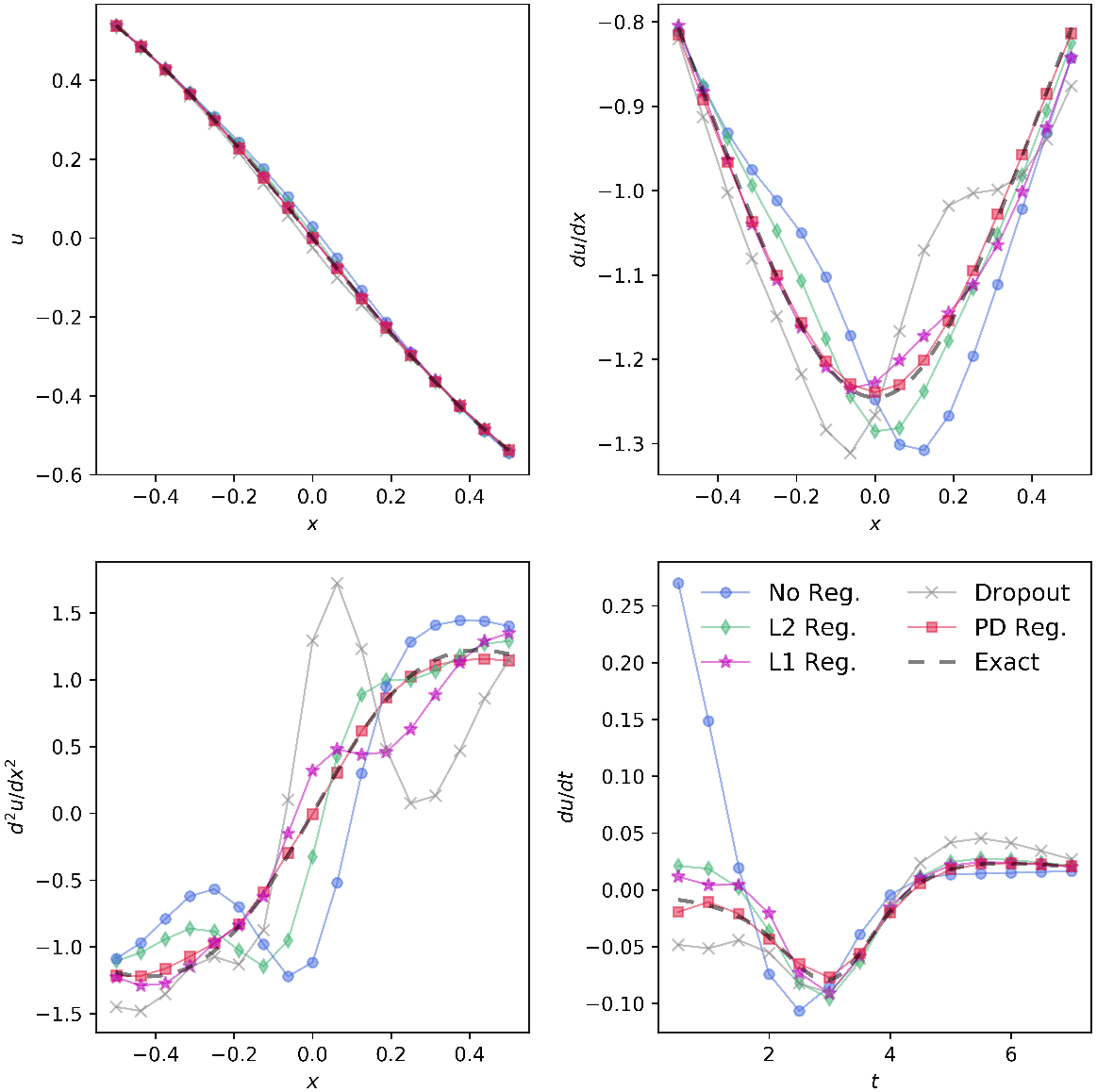}
		\caption{Comparison between the performance of different regularization methods in accurate prediction of first and second-order derivatives of the solution to the Burgers' equation. The solution and it's spatial derivatives are depicted at $t=2s$. The temporal derivatives are depicted at $x=0.04$.} 
		\label{Burgers_derivatives}
	\end{center}
\end{figure}

Figure \ref{Burgers_performance} shows a comparison between the performance of each of the regularization methods for different noise levels. For each regularization method, we train three different models, each trained using a different random selection of training data, and the results for each of the models as well as the average results are shown. It is evident that the proposed regularization method provides superior accuracies compared to the other methods at all the noise levels. Furthermore, Figure \ref{Burgers_derivatives} represents a comparison between the performance of different regularization methods in accurate prediction of first and second-order derivatives of the solution to the Burgers' equation with $\gamma$ set to 0. As can be seen,  all the  regularization methods, except  the proposed one, fail to provide accurate derivative values. This is a remarkable feature of the proposed regularization method, producing physically-interpretable derivatives that can be accurately used in subsequent calculations such as sensitivity analysis.

\subsection{DNN regression on response measurements from Navier-Stokes equations}

\begin{figure}
	\begin{center}
		\includegraphics[width=0.6\linewidth]{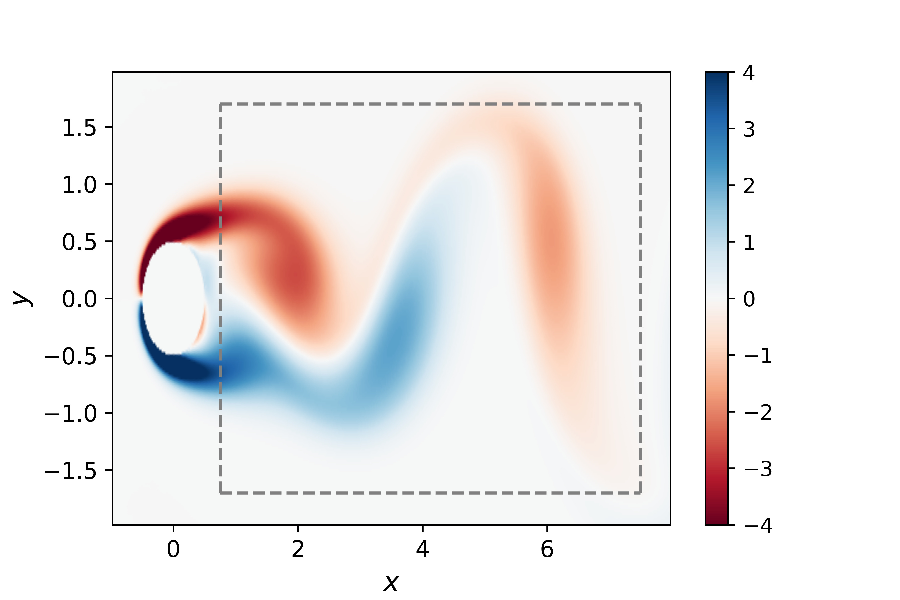}
		\caption{A snapshot of the vorticity field  $\omega$ obtained by solving the Navier-Stokes equations for the flow past a cylinder. The dashed gray box indicates the sub-sampling region (the figure and the sub-sampling region are adapted from \cite{raissi2018deep,raissi2019physics}).} 
		\label{NS_snapshot}
	\end{center}
\end{figure}

Let us now consider a system governed by the Navier-Stokes equations, and that  given a number of available response measurements our task is to construct a model that best fits to these measurements. Let us further assume that we are only aware of the fact that the system is governed by the Navier-Stokes equation, and have full knowledge about the parameters of the differential equation, but not about its  initial and boundary conditions. Similar to previous section, we generate synthetic response measurements by solving the following vorticity equation \cite{medjo1995vorticity}
\begin{equation}
\frac{\partial \omega}{\partial t}=-u\frac{\partial \omega}{\partial x}-v\frac{\partial \omega}{\partial y}+0.01\left( \frac{\partial^2 \omega}{\partial x^2}+\frac{\partial^2 \omega}{\partial y^2} \right),
\end{equation}
where $u$ and $v$ are respectively the $x$- and $y$-component of the velocity field, and $\omega$ is the vorticity, defined to be the curl of the velocity vector. In order to generate training, evaluation, and test datasets, we use the data provided by \cite{raissi2018deep}, which has been generated with the following specifications. An Immersed Boundary Projection Method \cite{taira2007immersed,colonius2008fast} is used to simulate the 2D fluid flow past a circular cylinder at Reynolds number $Re=100$. A multi-domain scheme~\cite{kutz2016dynamic} with four nested domains is used, with each successive grid twice as large as the previous one. Time and length are nondimensionalized. The flow has unit velocity and the cylinder has unit diameter. Data is collected on the highest-resolution domain with dimensions $9 \times 4$ with a resolution of $449 \times 199$. The Navier-Stokes solver uses a 3rd-order Runge-Kutta (RK3) scheme with time steps $\text{d}t=0.02$. Once the simulation converges to steady periodic vortex shedding, 151 flow snapshots are saved at each time step, out of which one is shown in figure \ref{NS_snapshot}. A small portion of the resulting data set is then sub-sampled to be used as the synthetic measurement data for the construction of our DNN models. Specifically, we subsample 5000 data points for training purposes. This figure and the sub-sampling region are adapted from \cite{raissi2018deep,raissi2019physics}.

Again in this example we construct five different models with five different regularization choices: no regularization, $L^2$ regularization, $L^1$ regularization, dropout, and the proposed regularization. The architecture of these DNN models is fixed and consists of 4 hidden layers, each with 128 units with Tanh nonlinearities. The output of this model is 3-dimensional, consisting of $\tilde{\omega}$, $\tilde{u}$, and $\tilde{v}$. The Adam optimization algorithm \cite{kingma2014adam} is used to solve the optimization problem defined in Equation \ref{minimize_loss}. The parameters $\beta_1$, $\beta_2$, and $\epsilon$ for the Adam optimizer are set to 0.9, 0.999, and $10^{-8}$, respectively. The batch size is set to 50. For the PD regularization, the following penalty term is used
\begin{equation}
J_{\mathcal{N}}\left(\bm{\Theta}; \bm{X}\right)= \frac{1}{2n}\sum_{i=1}^{n}\biggl[\frac{\partial \tilde{\omega}}{\partial t} \left(\bm{x}_i\right)+\tilde{u} \left(\bm{x}_i\right)\frac{\partial \tilde{\omega}}{\partial x} \left(\bm{x}_i\right)+  \tilde{v} \left(\bm{x}_i\right)\frac{\partial \tilde{\omega}}{\partial y} \left(\bm{x}_i\right) -0.01\left( \frac{\partial^2 \tilde{\omega}}{\partial x^2} \left(\bm{x}_i\right)+\frac{\partial^2 \tilde{\omega}}{\partial y^2} \left(\bm{x}_i\right)\right)\biggr] ^2,
\end{equation}
with the partial derivatives calculated using automatic differentiation.

Table \ref{table: example2} shows the hyperparameters that are tuned for each of the models, together with the search domain for each of the hyperparameters. Training is performed for 9 different number of epochs starting from 5,000 epochs and ending with 45,000 epochs. For each regularization method, given the number of epochs, we train 100 models on the training dataset. The model which results in the lowest relative $L_2$ norm on the evaluation dataset is then selected as the best model for the given number of training epochs. Once again, In order to eliminate the dependency of the reported relative $ L_2$ error norms to the selection of evaluation and test datasets, we choose to use a rather large number of our available synthetic data, and therefore we use 15000 randomly selected samples for evaluation and test purposes. 

\begin{table*}
	\begin{center}
		\caption{The hyperparameters that are tuned for each of the models together with their search domain.}
		\label{table: example2}
		\scalebox{0.74}{
			\begin{tabular}{|M{2.85cm}||M{2.95cm}|M{2.95cm}|M{2.95cm}|M{2.95cm}|M{2.95cm}|}
				\hline
				Regularization method & No Reg. & $L^2$ Reg. & $L^1$ Reg. & Dropout & PD Reg. \\ \hline
				Hyperparameters & $\eta \in [10^{-7},10^{-4}] $ & $\eta \in [10^{-7},10^{-4}] $ $\lambda_2 \in [10^{-5},10^{-1}]$ & $\eta \in [10^{-7},10^{-4}] $ $\lambda_1 \in [10^{-5},10^{-1}]$ & $\eta \in [10^{-7},10^{-4}] $ $P \in [0.9,0.999]$ & $\eta \in [10^{-7},10^{-4}] $ $\lambda_{\mathcal{N}} \in [10^{-3},10^{1}]$                 \\ \hline
		\end{tabular}}
	\end{center}
\end{table*}

\begin{figure}
	\begin{center}
		\includegraphics[width=.45\linewidth]{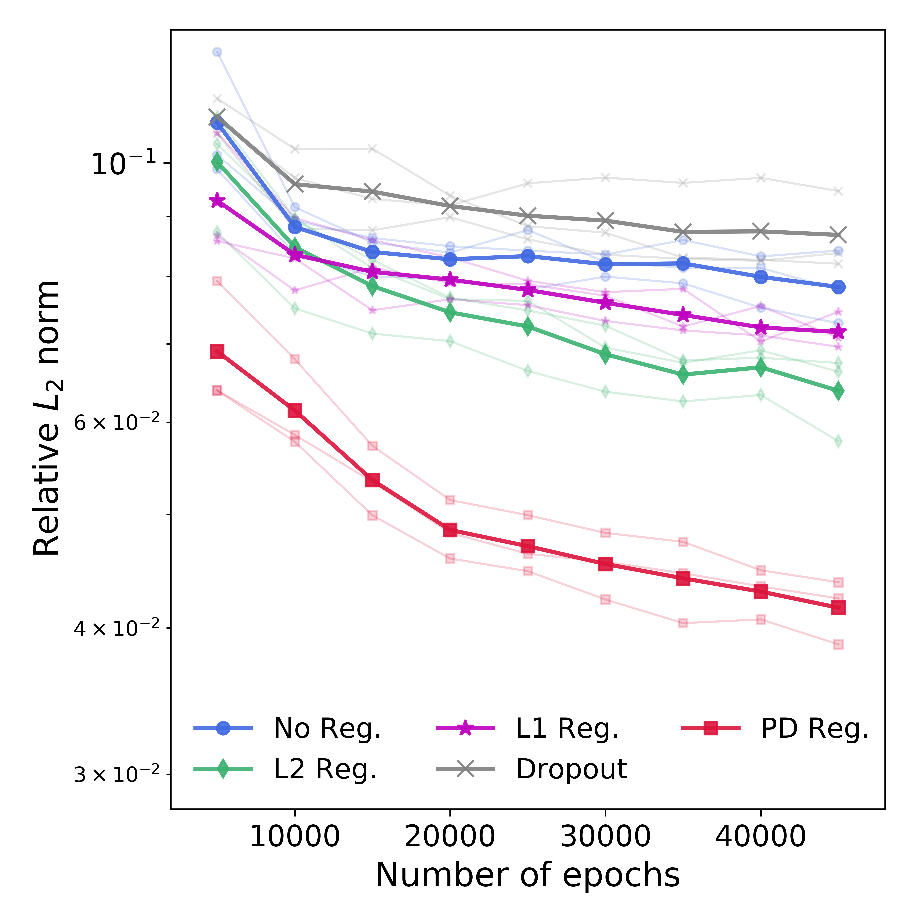}
		\centering	
		\caption{Comparison between the performance of different regularization methods in terms of relative $L_2$ norm for the Navier-Stokes example.} 
		\label{NS_performance}
	\end{center}
\end{figure}

Figure \ref{NS_performance} shows a comparison between the performance of different regularization methods. For each regularization method, we train three different models, each trained using a different random selection of training, evaluation, and test datasets, and the results for each of the models as well as the average results are shown. From this figure it is evident that the proposed regularization method provides superior generalization accuracies compared to other methods.

\subsubsection{Note on the poor performance of dropout} It is observed through the first two numerical examples that models trained with dropout have inferior accuracies compared to those trained with no regularization. Similar observation has been previously reported in  other studies, e.g. \cite{kingma2014adam}. There are multiple reasons that can explain this observation. Firstly, the success of dropout regularization has been mainly shown in the literature on classification tasks rather than on regression tasks, and also for Convolutional Neural Networks (CNN) rather than fully-connected DNNs \cite{srivastava2014dropout,krizhevsky2012imagenet,goodfellow2016deep}. Also, as stated earlier, at test stage a single unthinned network is used by implementing a weight scaling rule. However, the weight scaling rule is only an approximation for DNN models. It is only empirically shown (mostly on CNNs) that weight scaling rule performs well, and this has not been theoretically studied \cite{goodfellow2016deep}. It is stated in \cite{goodfellow2016deep} that the optimal choice of inference approximation for dropout networks is problem dependent, and weight scaling rule does not necessarily perform well generally for all the problems. Finally, dropout networks, compared to networks with no regularization, are known to require a relatively larger number of units/layers, and are required to be trained for a relatively larger number of epochs \cite{kingma2014adam}. However, this doesn't apply to our examples, where the network architecture and  number of training epochs are kept the same for all  the models. 

\subsection{Metamodeling for physical systems with random initial and boundary conditions}
In the first two examples, we demonstrated the effectiveness of the proposed regularization method in improving the generalization accuracy and interpretability of DNN models for measured data. In this and next example, we will implement the proposed regularization for constructing accurate metamodels for engineering systems. A metamodel serves as an approximating model for a quantity of interest (QoI), denoted by $u( \bm{x}  )$, especially when the QoI cannot be easily computed or measured.  The metamodel  is built to calculate the approximate QoI $\tilde{u}( \bm{x})$ by using  a set of $m$ exact model evaluations  $u (\bm{x}_{i}  )$    at the $d$-dimensional input locations $\{\bm{x}_{1}, \cdots, \bm{x}_{m}  \}$.

\begin{figure}
	\begin{center}
		\includegraphics[width=.5\linewidth]{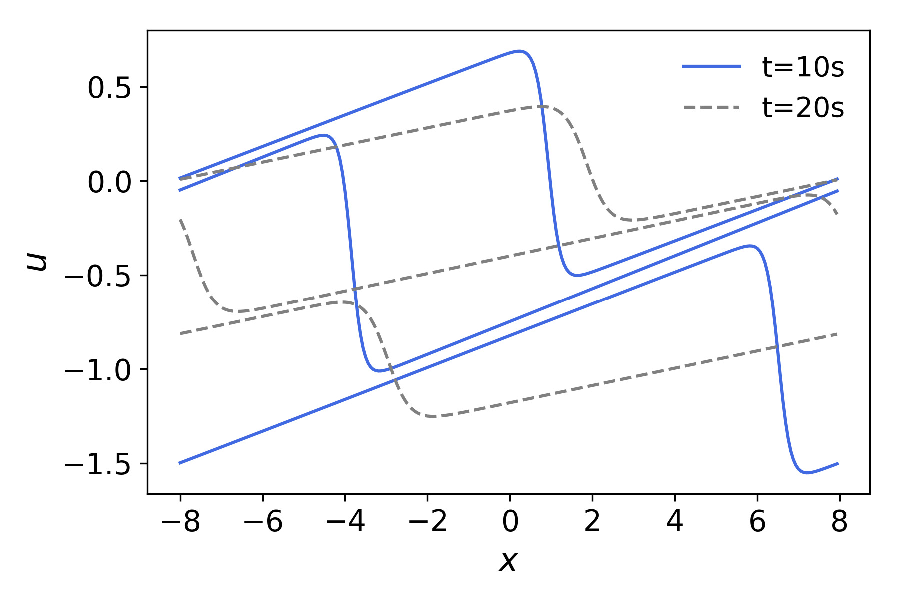}
		\centering	
		\caption{Sample responses of the Burgers' system calculated at random initial condition (Equation (\ref{random_initial})), and at $t=10s$  (blue solid lines), and $t = 20s$  (gray dashed lines).}
		\label{realizations}
	\end{center}
\end{figure}

Let us revisit the Burgers' system defined in Equation (\ref{Eqn:Burgers}). Here we assume the initial conditions takes the following random form:

\begin{equation}\label{random_initial}
u(0,x)=(-2+\xi_1)\sin(\pi x/8)+2\xi_2-1,
\end{equation}
where $\xi_1, \xi_2$ are i.i.d. random variables uniformly distributed in the range $[0,1]$. We seek to construct a DNN metamodel that can be used to calculate the response at any time $t \in [0,20]$ and any point $x \in [-8,8]$ , and for any given initial conditions. In order to generate the training data, we perform 200 simulations each with a  different initial conditions sampled from Equation~(\ref{random_initial}), according to the same numerical setup explained in Section \ref{example1}. A number of calculated system responses  at $t=10s, 20s$ are depicted in Figure \ref{realizations}. 

\begin{figure}
	\begin{center}
		\includegraphics[width=.5\linewidth]{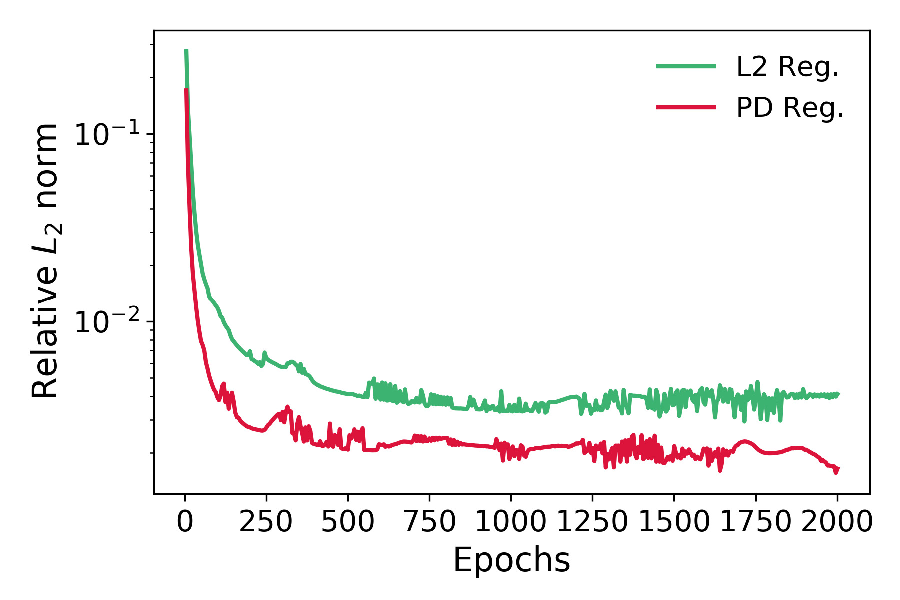}
		\centering	
		\caption{Comparison between the performance of PD and $L^2$ regularizations in terms of relative $L_2$ norm  for the Burgers' system with random initial conditions defined in Equation \ref{random_initial}.}
		\label{RBurgers_performance}
	\end{center}
\end{figure}

For the sake of brevity, we compare  our proposed regularization with  $L^2$ regularization, as it was shown, in the first example, to perform better than Dropout and $L^1$ regularizations. The inputs to our DNN metamodel are realizations of $t$, $x$, and $\xi_1, \xi_2$, and the output is the predicted scalar-valued response  $\tilde{u}$. The DNN architecture  consists of 7 hidden layers, each with 64 units with Tanh nonlinearities. The Adam optimization algorithm is used to solve the optimization problem defined in Equation (\ref{minimize_loss}). Parameters $\beta_1$, $\beta_2$, and $\epsilon$ for the Adam optimizer are set to 0.9, 0.999, and $10^{-8}$, respectively. Batch size is set to 100. For the PD regularization, the penalty term defined in Equation (\ref{Burgers_penalty}) is used. From the pool of data points obtained from our numerical simulations, we reserve 10\% for validation and 10\% for test purposes. For each of the two regularization method, we train 10 different metamodels, each for 2000 epochs, and select the one which results in the best validation accuracy. As can be seen in Figure \ref{RBurgers_performance},  physics-driven regularization  provides superior generalization accuracy compared to  $L^2$ regularization, showing the capability of the proposed PD regularization to also handle systems with known but uncertain initial conditions.

\subsection{Metamodeling in CFD-based design optimization}\label{example3}
In aerodynamics analysis and design problems, fluid flow is simulated by Computational Fluid Dynamics (CFD) solvers. This is done by solving the Navier-Stokes equations, which consist of mass and momentum conservation equations \cite{tu2018computational,nabian2016multiphase}. In the Eulerian framework, for 2D steady laminar flows, the mass conservation equation is given by
\begin{equation}\label{mass}
\nabla \cdot \bm{u} = 0,
\end{equation}
and the momentum conservation equation is defined as
\begin{equation}\label{momentum}
\bm{u} \cdot \nabla \bm{u} = - \frac{\nabla P}{\rho}+\nu \nabla ^2 \bm{u},
\end{equation}
where $\bm{u}=(u,v)$ is the 2D velocity vector, $P$ is the pressure, $\rho$ is the fluid density, and  $\nu$ is the fluid viscosity.

CFD simulation is known to be typically intensive in terms of computational time and memory usage. This  could potentially make a  CFD-based design space exploration prohibitively costly. As an alternative, metamodels can serve as substitutes  for fast fluid flow prediction and enable engineers and designers to perform design space exploration efficiently, especially at the early stages of design optimization when there is no need for high-fidelity simulations. 

In this example, we show how a PD-regularized Convolutional Neural Network (CNN) metamodel can produce accurate aerodynamic predictions towards accelerated design optimization. Specifically, we reconstruct the CNN model which was proposed in \cite{guo2016convolutional} and implemented in \cite{hennigh2018git} for the prediction of velocity field in 2D non-uniform steady laminar flows in the presence of rigid bodies, and will show how the  application of the proposed regularization to the CNN loss function can lead to  accuracy improvement, when compared to the state of the practice, as reported in \cite{hennigh2018git}. For a fair comparison, we use the same implementation as \cite{hennigh2018git}, with no changes to the datasets, network architecture, hyperparameters, and the number of training epochs. The only changes we made were  pertinent to  training the regularization.

In the past three examples, we regularized models in the form of feed-forward fully-connected DNNs, a form briefly explained in Section \ref{sec:Reg}. In this part we consider CNNs, which are a specific type of feed-forward DNNs. A CNN consists of recursive application of convolution and pooling layers, followed by fully-connected layers at the end of the network (as described in Section \ref{sec:Reg}). A convolution layer is a linear transformation that preserves spatial information in the input data. Pooling layers then simply reduce the dimensionality of the output of a convolution layer. More discussion on the CNNs can be found in \cite{lecun2015deep,goodfellow2016deep}. 

The training and validation datasets of \cite{hennigh2018git,guo2016convolutional}, used in this example, consist of five different types of simple geometric primitives, including triangles, quadrilaterals, pentagons, hexagons and dodecagons. Each sample is projected into a $256 \times 128$ Cartesian grid. The test dataset consists of different kinds of car prototypes including SUVs, vans, and sport cars. A binary representation of the geometry shapes is used, where a grid value is 1 if and only if it is within or on the boundaries of the geometry shapes, and a grid value is 0 otherwise. Each sample $i$ consists of five matrices $\{\bm{X_1}^{(i)},\bm{X_2}^{(i)},\bm{X_3}^{(i)},\bm{Y_1}^{(i)},\bm{Y_2}^{(i)}\}$ each of size $256 \times 128$, where the first matrix represents the geometry shape and the second and third matrices represent the x and y-components of the Cartesian grid, respectively. The latter two matrices represent the \emph{ground-truth} values for the x- and y-components of the velocity field, respectively, computed using the Lattice Boltzmann Method (LBM) \cite{chen1998lattice}. In all the experiments, the Reynolds number is set to 20. The no-slip boundary condition is applied to the geometry shape boundaries and horizontal walls. The training dataset contains 3,000 samples (600 samples for each type of primitives that are different in shape, size, location, and orientation). The validation dataset consists of 300 samples (60 different samples for each type of primitives). Finally, the test dataset consists of 28 car prototypes. The primitives and car prototypes in the training, validation, and test datasets are connected to the lower boundary.

The loss function used in \cite{guo2016convolutional,hennigh2018git} is in the form of
\begin{equation} \label{simple loss}
J\left(\bm{\Theta}; \bm{X},\bm{Y}\right)= \frac{1}{2n}\sum_{i=1}^{n} \biggl[ \left(\tilde{u} \left( \bm{X}^{(i)} \right) \otimes \mathbbm{1} \left( \bm{X_1}^{(i)}=0 \right) - \bm{Y_1}^{(i)} \right) ^2 + \left(\tilde{v} \left( \bm{X}^{(i)} \right) \otimes \mathbbm{1} \left( \bm{X_1}^{(i)}=0 \right) - \bm{Y_2}^{(i)} \right) ^2 \biggr],
\end{equation}
where $\bm{X}=\{\bm{X_1}^{(1)},\bm{X_2}^{(1)},\bm{X_3}^{(1)}, \cdots, \bm{X_1}^{(n)},\bm{X_2}^{(n)},\bm{X_3}^{(n)} \}$, $\bm{Y}=\{\bm{Y_1}^{(1)},\bm{Y_2}^{(1)}, \cdots, \bm{Y_1}^{(n)},\bm{Y_2}^{(n)}\}$, $\bm{X^{(i)}}=\{\bm{X_1}^{(i)},\bm{X_2}^{(i)},\bm{X_3}^{(i)} \}$, $n$ is the number of samples, $\mathbbm{1}$ is an indicator function and $\mathbbm{1} \left( \bm{X_1}^{(i)}=0 \right)$ has the same size as $\bm{X_1}^{(i)}$, $\tilde{u} \left( \bm{X}^{(i)} \right)$ and $\tilde{v} \left( \bm{X}^{(i)} \right)$ are the CNN predictions for the x and y-components of the velocity field for the sample $i$. The loss function in  Equation (\ref{simple loss}) is simply the Euclidean loss function that takes into account only the fluid part of the computational domain. As the competitor for our proposed method, we consider the metamodel trained with dropout with  $P=0.7$.

In order to do apply the proposed regularization, we only add a penalty term for the violation of the divergence-free condition (i.e. Equation (\ref{mass})) of the velocity field, and leave violations from the momentum conservation equation (Equation (\ref{momentum})) unpenalized. This is because the second penalty term would necessitate another metamodel to be built for the pressure field $P$. Therefore, since in the competing study \cite{hennigh2018git},  no metamodel for the pressure field was constructed, for the sake of a fair comparison, we only applied regularization to the velocity field metamodel. As a result, the PD regularization is given by
\begin{equation}\label{PI Example3}
J_{\mathcal{N}} \left(\bm{\Theta}; \bm{X}\right)= \frac{1}{2n}\sum_{i=1}^{n} \biggl[ \frac{\partial \tilde{u}}{\partial x} \left( \bm{X}^{(i)} \right) \otimes \mathbbm{1} \left( \bm{X_1}^{(i)}=0 \right) + \frac{\partial \tilde{v}}{\partial y} \left( \bm{X}^{(i)} \right) \otimes \mathbbm{1} \left( \bm{X_1}^{(i)}=0 \right) \biggr]^2,
\end{equation}
and the partial derivatives are calculated using automatic differentiation. It should be noted that we are incorporating only partial prior knowledge about the physics  in the regularization, and potential further improvement can be expected with the inclusion of the momentum penalty term.   

For the CNN metamodel a U-network approach is used with residual layers \cite{he2016deep} similar to Pixel-CNN++ \cite{van2016conditional,Salimans2017PixeCNN} which is a class of powerful generative models \cite{goodfellow2016nips}. For implementation, we used the source code provided by \cite{hennigh2018git}. The Adam optimization algorithm is used to solve the optimization problem defined in Equation \ref{minimize_loss}. Parameters $\beta_1$, $\beta_2$, and $\epsilon$ for the Adam optimizer are set to 0.9, 0.999, and $10^{-8}$, respectively. Batch size is set to 8. Learning rate is set to $10^{-4}$. For the PD regularization, the hyperparameter $\lambda_{\mathcal{N}}$ is tuned  by training 5 different models and selecting the best one using cross-validation.

\begin{figure}
	\begin{center}
		\includegraphics[width=.65\linewidth]{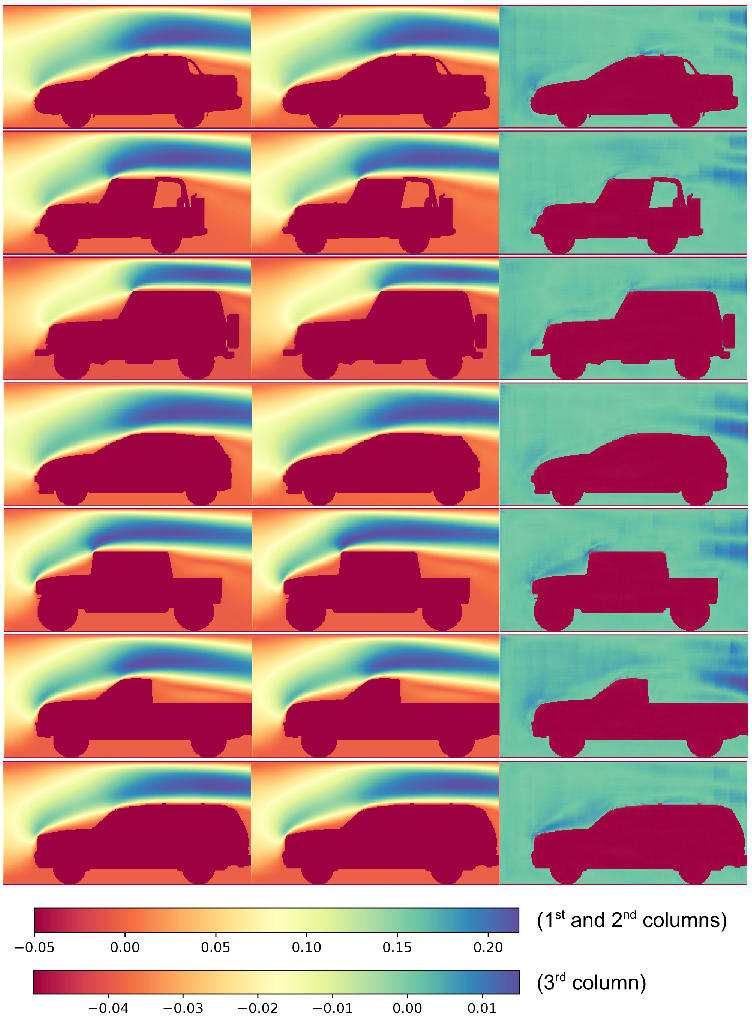}
		\centering	
		\captionsetup{}
		\caption{Visualization of the velocity magnitude. The first column shows the LBM ground truth results. The second column shows the CNN prediction results using the proposed PD regularization method. The third column shows the magnitude of the difference between ground truth and predicted results. The car bodies are shown with a magnitude of -0.05.} 
		\label{Cars}
	\end{center}
\end{figure}

Figure \ref{Cars} shows a visualization of the velocity field for the test data. The first column shows the LBM ground truth results. The second column shows the CNN prediction results using the proposed regularization method. The third column shows the  $L_2$ norm of the difference between ground truth and predicted results, which are averaged over three independent training efforts. It is evident that the results are in close agreement. Table \ref{table: example3} shows a comparison between the performance of the metamodels trained with different regularization methods. The third column shows the state of the practice \cite{hennigh2018git}. It can be seen that dropout regularization (third column) and PD regularization (fourth column) have similar performances. It should be noted again that the applied PD regularization only  incorporates  the partial prior knowledge  pertaining to the mass conservation, and doesn't regularize based on the momentum equation. However, it is observed that the best performance is obtained when PD regularization is applied in addition to dropout. Specifically, the application of our proposed regularization method to the dropout implementation of \cite{hennigh2018git} has reduced the relative $L_2$ norm by $16.46\%$.

\begin{table*}
	\begin{center}
		\caption{A comparison between the performance of metamodels trained with different regularization methods. }
		\label{table: example3}
		\scalebox{0.85}{
			\begin{tabular}{|M{2.5cm}||M{2.5cm}|M{2.5cm}|M{2.5cm}|M{3.8cm}|}
				\hline
				Regularization method & No Reg. & Dropout ($P=0.7$) & PD Reg. ($\lambda_{\mathcal{N}}=0.1$) & Dropout ($P=0.7$) \&~PD Reg. ($\lambda_{\mathcal{N}}=0.1$) \\ \hline
				Relative $L_2$ norm & $2.55 \times 10^{-2}$ & $2.43 \times 10^{-2}$ & $2.45 \times 10^{-2}$ & $2.03 \times 10^{-2}$\\ \hline
		\end{tabular}}
	\end{center}
\end{table*}

\section{Conclusion}\label{sec:conclusion}
In this work, we presented a physics-driven regularization method for training of DNN models and metamodels. It has been shown through four numerical examples (systems governed by the Burgers' and Navier-Stokes equations) that the proposed regularization method results in models and metamodels that are physically interpretable, and  compared to other common regularization methods, it results in superior generalization accuracies, which can potentially  enhance engineering design and analysis such as aerodynamic design optimization of passenger vehicles, as discussed in the last example. This is achieved by applying a regularization term to the optimization loss function that discourages the models and metamodel from violating the governing laws. 

The computational cost associated with PD regularization varies depending on the underlying physics. Specifically, using automatic differentiation, calculating first-order derivatives of DNN outputs with respect to inputs  is computationally cheap and even negligible, as it only needs  minor modifications to the derivative information already available from backpropagation. For differential equations with higher-order terms, on the other hand,  additional derivative calculations incur additional computational costs. However, this extra computational time is typically allowable   since (1) in data-driven modeling of physical systems (e.g. Examples 1 and 2), generalization accuracy and interpretability may be more important factors, as is reflected in the wide use of deep neural networks which are computationally more expensive than other machine learning methods; and (2) in data-driven metamodeling problems (e.g. Examples 3 and 4), the computational cost associated with training data generation is significantly larger than the training cost.  With these in mind, in order to reduce the extra cost, instead of automatic differentiation, one can use the Monte Carlo approach introduced  in \cite{sirignano2018dgm}, or  the first-order variational form of differential equations, as proposed in \cite{nabian2018deep2}.

One limitation of the proposed regularization method is that, in order for us to promote the DNNs that satisfy the governing equations, an independent model has to be constructed for each of the physical variables that appear in the governing laws. This was discussed in Section \ref{example3} where a separate  `pressure field' metamodel was needed to enforce the momentum conservation. In the presence of available labeled training data for all the physical variables that appear in the governing laws, we can construct separate models for each physical variable in order to enable the PD regularization. This can be done even if some of those physical variables are not our QoIs.  This will in fact increase the computational cost compared to other regularization alternatives, which only account for the QoIs. But, a different investigation can determine  whether the accuracy improvement is worth the the extra cost.

There exists a series of research opportunities to pursue in  future studies as extensions pertinent to this work, including the following: (1) The performance of hybrid regularization techniques that make apply PD regularization together with other regularization methods, such as $L^2$, $L^1$, and/or dropout, can be investigated further to evaluate whether the test accuracy gains are accumulated. A glimpse of this hybrid use is already discussed in Section \ref{example3}, however, more comprehensive studies are needed; (2) It is worthwhile  to comparatively study the PD regularization  with respect to other regularization techniques such as regularizations based on Lipschitz continuity (e.g. \cite{gouk2018regularisation,gulrajani2017improved,miyato2018spectral}), Jacobian regularization (e.g. \cite{gu2014towards}), or adversarial training \cite{goodfellow2014explaining}. Notably, regularizations based on Lipschitz continuity have shown great success in improving the stability and robustness of the Generative Adversarial Networks \cite{gulrajani2017improved,miyato2018spectral}. (3) In order to enhance the training, It is worthwhile to investigate optimal sampling strategies to intelligently select the training data in order to reduce the DNN generalization error and also improve convergence rate for the proposed regularization method; (4) One can also  study the performance of the proposed method in constructing accurate models and metamodels for nonlinear dynamic systems with varying system parameters, for which some variations of the system parameters may lead the system to go through bifurcations (such as a system with varying Reynolds number that is governed by the Navier-Stokes equations); and finally (5)  It is worthwhile to study when in the training phase the regularizing term is applied. In the current algorithm, the regularizer  $J_{\mathcal{N}}$ is added to the loss function (Equation \ref{PI loss}) from the beginning of training phase. Future study can determine whether rate of convergence can be improved if this is done after a few iterations in  the training phase.

\bibliography{bibfile}

\end{document}